%% file: WangTongZhao24CISS.tex
\documentclass[conference]{IEEEtran}
\IEEEoverridecommandlockouts
\usepackage{cite}
\usepackage{hyperref}       
\usepackage{url}            
\usepackage{booktabs}       
\usepackage{amsfonts}       
\usepackage{nicefrac}       
\usepackage{microtype}      
\usepackage{xcolor}         
\usepackage{amsmath}
\usepackage{graphicx}
\usepackage{subcaption}
\usepackage{multirow}
\usepackage{mathrsfs}
\usepackage{bbm}
\usepackage{xr}
\usepackage{bm}
\usepackage{makecell}
\def\BibTeX{{\rm B\kern-.05em{\sc i\kern-.025em b}\kern-.08em
    T\kern-.1667em\lower.7ex\hbox{E}\kern-.125emX}}

\input{LTmacros}

\def\xbf{{\bm x}}
\def\Xbf{{\bm X}}
\begin{document}

\title{Generative Probabilistic Time Series Forecasting and Applications in Grid Operations
\thanks{This work was supported in part by the National Science Foundation under Award 2218110.}}

\author{\IEEEauthorblockN{Xinyi Wang}
\IEEEauthorblockA{\textit{Electrical and Computer Engineering} \\
\textit{Cornell University}\\
Ithaca, USA \\
xw555@cornell.edu\\}
\and
\IEEEauthorblockN{Lang Tong}
\IEEEauthorblockA{\textit{Electrical and Computer Engineering} \\
\textit{Cornell University}\\
Ithaca, USA \\
lt35@cornell.edu}
\and
\IEEEauthorblockN{Qing Zhao}
\IEEEauthorblockA{\textit{Electrical and Computer Engineering} \\
\textit{Cornell University}\\
Ithaca, USA \\
qz16@cornell.edu}}

\maketitle

\begin{abstract}
Generative probabilistic forecasting produces future time series samples according to the conditional probability distribution given past time series observations. Such techniques are essential in risk-based decision-making and planning under uncertainty with broad applications in grid operations, including electricity price forecasting, risk-based economic dispatch, and stochastic optimizations. Inspired by Wiener and Kallianpur's innovation representation, we propose a weak innovation autoencoder architecture and a learning algorithm to extract independent and identically distributed innovation sequences from nonparametric stationary time series.  We show that the weak innovation sequence is Bayesian sufficient, which makes the proposed weak innovation autoencoder a canonical architecture for generative probabilistic forecasting. The proposed technique is applied to forecasting highly volatile real-time electricity prices, demonstrating superior performance across multiple forecasting measures over leading probabilistic and point forecasting techniques.
\end{abstract}


\section{Introduction}
\label{sec:intro}
\input{intro_final}

\section{Innovation Representation}
\label{sec:innovations}
\input{representation_final}

\section{Generative Probabilistic Forecasting}
\label{sec:forecasting}
\input{forecasting_final}

\section{Learning Innovation Representation}
\label{sec:wiae}
\input{WIAE_training_final}

\section{Numerical Results}
\label{sec:simulation}
\input{simulation_final}

\section{Conclusion}
This paper presents a novel approach to generative probabilistic forecasting (GPF) of nonparametric time series with unknown probability structures, reviving the classic idea of innovation representation of Wiener-Kallianpur and Rosenblatt with modern generative machine learning techniques. Our main theoretical contribution lies in establishing that the weak innovation process of a time series forms Bayesian sufficient statistics for probabilistic forecasting. These results imply that WIAE is a canonical architecture for stochastic decision-making such as probabilistic forecasting.  We have also made a methodological advance in developing a  generative forecasting algorithm that produces Monte Carlo samples of future time series realizations. The application of the proposed technique to real world datasets demonstrates that innovation-based forecasting offers superior performance across multiple performance measures over leading benchmarks.

The vanilla implementation of GPF presented here can be extended in multiple directions; chief among them is the multivariate GPF. While the underlying principle of innovation-based forecasting applies to multivariate time series, the condition under which weak innovation representation exists may become more tenuous and the training of WIAE more challenging. More sophisticated neural network implementations such as RNN and LSTM may also improve the forecasting performance.

\bibliography{ts_forecast}
\bibliographystyle{IEEEtran}

\end{document}


%

%

\onecolumn
\aistatstitle{Non-parametric Probabilistic Time Series Forecasting via Innovations Representation: \\
Supplementary Materials}

\input{appendix}

\bibliography{ts_forecast}












\vfill

%% file: LTmacros.tex
\newtheorem{theorem}{Theorem}

\def\beq{\begin{equation}}
\def\eeq{\end{equation}}
\def\bea{\begin{eqnarray}}
\def\eea{\end{eqnarray}}
\def\ba{\begin{array}}
\def\ea{\end{array}}

\def\bitem{\begin{itemize}}
\def\eitem{\end{itemize}}
\def\ben{\begin{enumerate}}
\def\een{\end{enumerate}}

\definecolor{bgrd}{rgb}{1,1,1}
\definecolor{gray}{rgb}{0.5,0.5,0.5}
\definecolor{dkr}{rgb}{0.7,0.1,0.2}
\definecolor{dkb}{rgb}{0.1,0.1,0.8}

\makeatletter
\newdimen{\captionwidth}
\long\def\@makecaption#1#2{%
\captionwidth .9\hsize
\vskip 10pt%
\setbox\@tempboxa\hbox{#1: #2}%
  \ifdim \wd\@tempboxa >\captionwidth%
    \setbox\@tempboxa\hbox{#1:\hspace*{.5em}}%
    \hfil\parbox{\captionwidth}{\raggedright\hangindent \wd\@tempboxa%
    \hangafter=1\unhbox\@tempboxa#2}\hfill%
  \else\centerline{\box\@tempboxa}%
  \fi
}
\makeatother

\newcommand{\mbbE}{\mathbb {E}}

\newcommand{\Xmsc}{\mathscr{X}}

\def\nubf{\hbox{\boldmath$\nu$\unboldmath}}

\def\xbf{{\bf x}}

\def\xbf{{\bf x}}

\def\Vbf{{\bf V}}

\def\Xbf{{\bf X}}

\def\Uc{{\cal U}}

%% file: intro_final.tex
We develop a {\em generative probabilistic forecasting (GPF) technique} for unknown and nonparametric time series models. Whereas standard probabilistic forecasting aims to estimate the conditional probability distribution of the time series at a future time, GPF obtains a generative model capable of producing arbitrarily many Monte Carlo samples of future time series realizations according to the conditional probability distribution of the time series given past observations.  As a nonparametric probabilistic forecasting technique, GPF is essential for decision-making under uncertainty where data-driven risk-sensitive optimization requires conditional samples of future randomness. The Monte Carlo samples generated from GPF can be used to produce any form of point forecast.

Nonparametric probabilistic time series forecasting is challenging because the joint probability distributions that characterize temporal dependencies are infinite-dimensional and unknown. Most approaches rely on parametric and semiparametric models, restricting the infinite-dimensional inference to a finite-dimensional parameter space. Classic examples include the forecasting based on autoregressive moving average, GARCH, and Gaussian process models \cite{weron_forecasting_2008,weron_electricity_2014,hardle_review_1997}.

This work pursues a new path to nonparametric probabilistic forecasting, following the generative AI principle and inspired by the classic notion of {\em innovation representation} pioneered by Norbert Wiener and Gopinath Kallianpur in 1958 \cite{Wiener:58Book}.
In the parlance of modern machine learning, the Wiener-Kallianpur innovation representation is an autoencoder, where the {\em causal encoder} transforms a stationary time series $(X_t)$ to an {\em innovation process} $(V_t)$ defined by the independent and identical (i.i.d.) uniformly distributed sequence, followed by a {\em causal decoder} that reproduces almost surely the original time series.
The most striking feature of the Wiener-Kallianpur innovation representation is that $V_t$ at time $t$ is statistically independent of the past $\Xmsc_{t}:=\{X_{t-1}, X_{t-2}, \cdots\}$, implying that $V_t$ contains only new information at time $t$. The autoencoder's capability to reproduce $(X_t)$ makes the latent process sufficient statistics in any decision-making. The innovation being i.i.d. uniform means that the autoencoder captures the complete model temporal dependencies of the time series. This particular feature of the latent process plays a critical role in the proposed GPF methodology.

The main challenge of applying Wiener-Kallianpur innovation representation for inference and decision-making is twofold. First, obtaining a causal encoder to extract the innovation process requires knowing the marginal and joint distributions of the time series, which is rarely possible without imposing some parametric structure. Furthermore, even when the probability distribution is known, there is no known computationally tractable way to construct the causal encoder to obtain an innovation process. Second, the Wiener-Kallianpur innovation representation may not exist for a broad class of random processes, including some of the important cases of finite-state Markov chains \cite{Rosenblatt:59}. These conceptual and computational barriers prevent employing innovation representation to a broad class of inference and decision problems except for Gaussian and additive Gaussian models~\cite{Kailath1968TAC}.

\subsection{Summary of Contributions}
This paper makes methodological and practical advances in nonparametric probabilistic time series forecasting in three aspects. First, we propose a Weak Innovation AutoEncoder (WIAE) and a deep learning algorithm based on Rosenblatt's weak innovation representation \cite{Rosenblatt:59}, a relaxation of Wiener-Kallianpur’s strong innovation representation, significantly extending the class of time series for which the canonical uniform i.i.d. innovation representation exists. We show further that the latent process of WIAE forms Bayesian sufficient statistics for generative probabilistic forecasting, resulting in no loss of optimality when the weak innovation representation is used.

Second, we propose Generative Probabilistic Forecasting with Weak Innovation (GPF-WI), a WIAE-based deep learning technique that produces future time series samples according to the conditional probability distribution on past realizations. Through WIAE, GPF-WI transforms the problem of generating Monte Carlo samples of the future time series with unknown probability distributions to using past innovation observations and Monte Carlo samples from the i.i.d. uniform distribution. See Sec.~\ref{subsec:inference}.

Third, we apply GPF-WI to probabilistic forecasting of highly volatile wholesale electricity prices using real datasets from several major independent system operators. The frequent price spikes due to increasing renewable integration and significant weather events make conventional techniques ineffective. The empirical results demonstrate marked improvement over existing solutions, scoring near the top across multiple datasets and under different performance metrics. See Sec.~\ref{sec:simulation}.

\subsection{A Contextual Review of Literature}
\label{subsec:literature}

Nonparametric probabilistic time series forecasting has a long history. See \cite{hardle_review_1997} for a review up to the late 1990's. Machine learning techniques gained prominence in the last decade, which were reviewed and compared by \cite{review_nonparametric}. Here we restrict our discussion to those machine learning techniques sharing architectural or methodological similarities with our approach. In particular, we focus on autoencoder-based forecasting techniques and the generative methods that rely on Monte-Carlo sampling to produce realizations of forecasts.

Most autoencoder-based GPF techniques rely on state-of-the-art autoencoder architectures for latent representation learning, followed by producing forecasts by Monte Carlo sampling of the latent representation. These methods generally adopt likelihood-based generative models such as Variational Autoencoder (VAE) \cite{nguyen_temporal_2021,li_synergetic_2021}, normalizing flows \cite{rasul_multivariate_2022} and denoising diffusion probabilistic models \cite{Li_2022_diffusion}. For instance, the authors in \cite{nguyen_temporal_2021} developed a technique that models the conditional distribution of future time series as a Gaussian latent process of an autoencoder trained with historical data. A normalizing flow model conditioned on Recurrent Neural Network (RNN) or transformer model is adopted to learn the conditional probability distribution \cite{rasul_multivariate_2022}.

Both VAE and diffusion learning rely on statistically independent training samples with log-likelihood function. When deriving the variational lower bound for log-likelihood and implementing it for training, those methods assume that the samples in the training set are independent, which simplifies the minimization of an upper bound of negative log-likelihood. For time series segments, the samples aren't independent. Hence, minimizing the joint log-likelihood of multiple samples is intractable, making these methods not suitable for time series. In contrast, when learning weak innovation representation through WIAE, as considered in our approach, time series training samples are allowed to have unknown temporal dependencies.

Novel learning architectures capable of modeling complex temporal dependencies have been proposed in recent years. In \cite{salinas_deepar_2019,wang_deep_2019,du_probabilistic_2022}, the authors adopt RNN to develop a variety of parametric probabilistic forecasting techniques. The authors in \cite{oord_wavenet_2016,borovykh_conditional_2018} adopted a dilated convolutional network to learn the conditional distribution.

The success of transformer-based deep learning in natural language processing inspired its applications to time series forecasting \cite{zhou_informer_2021,zhou_fedformer_2022,liu_pyraformer_2022}, some demonstrating promising performance. These point forecasting techniques have not been tested on the more challenging time series with high volatility, such as electricity prices. Although these techniques do not produce probabilistic forecasts, their strong performance in point forecasting makes them worth comparing with.

Clarification of notations used is in order. Random variables are highlighted in capital letters and their realizations in lower cases. When needed for clarity, vectors are in boldface. We use $(X_t)$ for a time series indexed by $t$, extending from $-\infty$ to $\infty$. For a partial time series realization up to time $t$ is denoted by $\xbf_t:=(x_t, x_{t-1},\cdots)$ and $\xbf_{t_o:t}:=(x_t, x_{t-1},\cdots, x_{t_o})$.

%% file: representation_final.tex
\begin{figure}[h]
    \centering
    \includegraphics[scale=0.6]{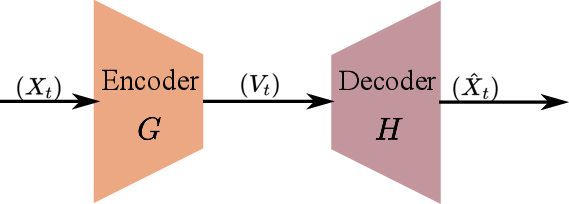}
    \caption{An autoencoder interpretation of innovation representation.}
    \label{fig:inn-ae}
    \vspace{-1em}
\end{figure}

\label{subsec:inn}
Innovation representation can be viewed as a causal autoencoder transformation $(G,H)$, as illustrated by Fig.~\ref{fig:inn-ae}. A key characteristics separating innovation autoencoder from other autoencoder architectures is that the latent process $(V_t)$ is an i.i.d. uniform sequence extracted  through the causal encoder $G$, and it can be used to produce an estimate $(\hat{X}_t)$ of $(X_t)$ by a causal decoder $H$.  Specifically,
\begin{subequations}
\renewcommand{\theequation}{\theparentequation.\arabic{equation}}
    \begin{align}
    V_t &= G(X_t,X_{t-1},\cdots), \label{Eq:encoder}\\
    \hat{X}_t &= H(V_t,V_{t-1},\cdots), \label{Eq:decoder}\\
    &V_t\stackrel{\mbox{\sf\tiny i.i.d.}}{\sim} \Uc[0,1].\label{Eq:iid}
\end{align}
\label{eq:wiae}
\end{subequations}
The temporal independence of $(V_t)$ implies that $V_t$ is independent of the past $\mathbf{X}_t:=(X_{t-1},X_{t-2},\cdots)$.
Hence, $V_t$ represents the new information not contained in $\mathbf{X}_t$.

The criterion for reconstruction distinguishes different types of innovation representations. The Wiener-Kallianpur innovation representation originally proposed by Wiener \cite{Wiener:58Book} requires almost sure equality between $(X_t)$ and $(\hat{X}_t)$, hence referred to as {\em strong innovations representation.} That is, the transformation pair $(G,H)$ is causally invertible ensuring that $(X_t)\stackrel{a.s.}{=}(X_t)$. The causal invertibility makes the Wiener-Kallianpur innovation sequence $(V_t)$ a sufficient statistic. Therefore, the decision-making problem based on strong innovations is lossless. Unforunately, the Wiener-Kallianpur innovation can only be analytically computed for stationary Gaussian time series \cite{Kalman:60TASME} and additive Gaussian models \cite{Kailath1968TAC}. For those time series, the prediction error of the nonlinear minimum mean squared error predictor is its innovation sequence. Methods of extracting innovations from nonparametric time series were mostly unknown until recently when obtained the first machine learning approach that learns the autoencoder of the Wiener-Kallianpur innovation representation \cite{WangTong:21JMLR}.

Rosenblatt was the first to state that the Wiener-Kallianpur innovation does not exist for a broad class of random processes~\cite{Rosenblatt:59}.
He thus suggested a weaker version of the Wiener-Kallianpur innovation representation by replacing the almost sure equality with distribution-wise equality:
\begin{align}
    (\hat{X}_t) \stackrel{d}{=} (X_t).  \label{Eq:recons}
\end{align}
Herein, we refer to the innovation representation with matching input-output in distribution as weak innovation representation and $(G,H)$ as the weak innovation autoencoder.

Although the existence of a WIAE remains an open problem, Rosenblatt has shown that his relaxation broadens the class of random processes for which weak innovation exists.  With $(V_t)$ not necessarily a sufficient statistic, decision-making based on $(V_t)$ is lossy in general. An open question is whether forecasting decisions based on a weak innovation representation perform optimally.  To this end, we provide a definitive answer for the probabilistic forecasting problem in Sec.~\ref{sec:forecasting}. Currently, there are no known techniques to extract weak innovation sequences from time series.  We propose the first such technique based generative adversary network (GAN) learning in Sec.~\ref{sec:wiae}.

%% file: forecasting_final.tex
Consider a stationary time series $(X_t)$.  The probabilistic forecasting of $X_{t+T}$ given current and past realizations $\Xbf_t=\xbf_t$ is to estimate the conditional distribution $F_{t+T|t}(x|\xbf_t)$:
\begin{align}
    F_{t+T|t}(x|\xbf_t):=\Pr[x_{t+T}\le x|\Xbf_t=\xbf_t].
    \label{eq:conditional_prob}
\end{align}
 In contrast, GPF produces Monte Carlo samples of $X_{t+T}$ with the conditional distribution $F_{t|t+T}$. This is nontrivial because of the unknown and complex temporal dependencies of $(X_t)$.
 
\subsection{GPF with Weak Innovation}
\label{subsec:inference}
\begin{figure}[h]
    \centering
    \includegraphics[width=\linewidth]{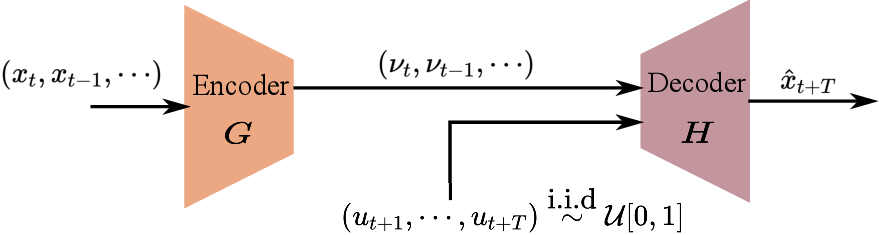}
    \caption{Probabilistic Forecasting via Weak Innovations.}
    \label{fig:forecasting}
    \vspace{-1em}
\end{figure}

This section introduces GPF-WI for forecasting. Generating samples directly from $F_{t+T|T}$ is difficult. As shown in Fig.~\ref{fig:forecasting},  GPF-WI generates conditional samples $(V_{t+T},\cdots, V_{t+1})$ from the i.i.d. uniform distribution, and uses WIAE decoder to produce future samples.  The key advantage of using innovations rather than the original time series is that future innovations are statistically independent of the past observations $\Vbf_t=\nubf_t$.

As illustrated by Fig.~\ref{fig:forecasting}, the weak innovations up to time $t$ is the output of the weak innovation encoder $G$. Although future innovation $\{V_{t+1},\cdots,V_{t+T}\}$ are not yet realized, the fact that they are i.i.d. uniform allows us to sample them from the uniform marginal.
These Monte Carlo samples $(u_{t+1},\cdots, u_{t+T})$, along with the encoder outputs $(\nu_t,\nu_{t-1},\cdots)$, form the input of the decoder $H$ that produces an estimate of $x_{t+T}$.
Different sets of Monte Carlo samples produce different estimates of $X_{t+T}$, generated from its conditional probability distribution.

While  GPF-WI is easy to implement once the WIAE has been learned, 
one should question whether replacing $\Xbf_t$ by $\Vbf_t$ is justified, since weak innovation sequence may not be a sufficient statistics. We address this question next.

\subsection{Weak Innovation are Bayesian Sufficient}
The objective of probabilistic forecasting is to estimate the conditional probability distribution, which is deterministic once past observation is given.  In contrast, GPF aims to obtain a generative model that produces estimates of a set of realizations with conditional $F_{t+T|t}$. In this aspect, GPF is a Bayesian estimation problem, where the notion of sufficient statistics takes a different form.  In particular, $T(X)$ is Bayesian sufficient for estimating a realization of random variable $Y$ if the posterior distribution given $X$ is the same given $T(X)$ \cite{BickelDoksum07:book}.

The theorem below establishes that the weak innovation sequence $\nubf_t=G(\xbf_t)$, under a mild assumption on WIAE, is Bayesian sufficient  for estimating realizations of $X_{t+T}$.
For simplicity, we define the following notation for the conditional distribution of $X_{t+T}|\Vbf_t$:
\[F^{(\nu)}_{t+T|t}(x|\nubf_t):= \Pr[V_{t+T}\leq x|\Vbf_t=\nubf_t].\]
\begin{theorem}[Bayesian Sufficiency]
Let $(X_t)$ be a stationary time series for which the weak innovation exists.
Let $(V_t)$ be the weak innovation sequence of $(X_t)$, and assume that the causal decoder $H$ is injective.
Then, for almost all $\xbf_t$ and $x$ (with respect to Lebesgue measure),
\[F_{t+T|t}(x|\xbf_t)=F_{t+T|t}^{(\nu)}(x|\nubf_t),\]
where $F_{t+T|t}(x|\xbf_t)$ is defined in \eqref{eq:conditional_prob}, $\nubf_t=G(\xbf_t)$, and
\[
F_{t+T|t}^{\nu}(x|\nubf_t):=\Pr(X_{t+T}\le x|\nubf_t).
\]
\label{thm:sufficiency}
\end{theorem}
\vspace{-1em}
{\em Proof:} See the appendix of \cite{wang2023nonparametric}.

%% file: WIAE_training_final.tex
\begin{figure}[t]
    \centering
    \includegraphics[width=\linewidth]{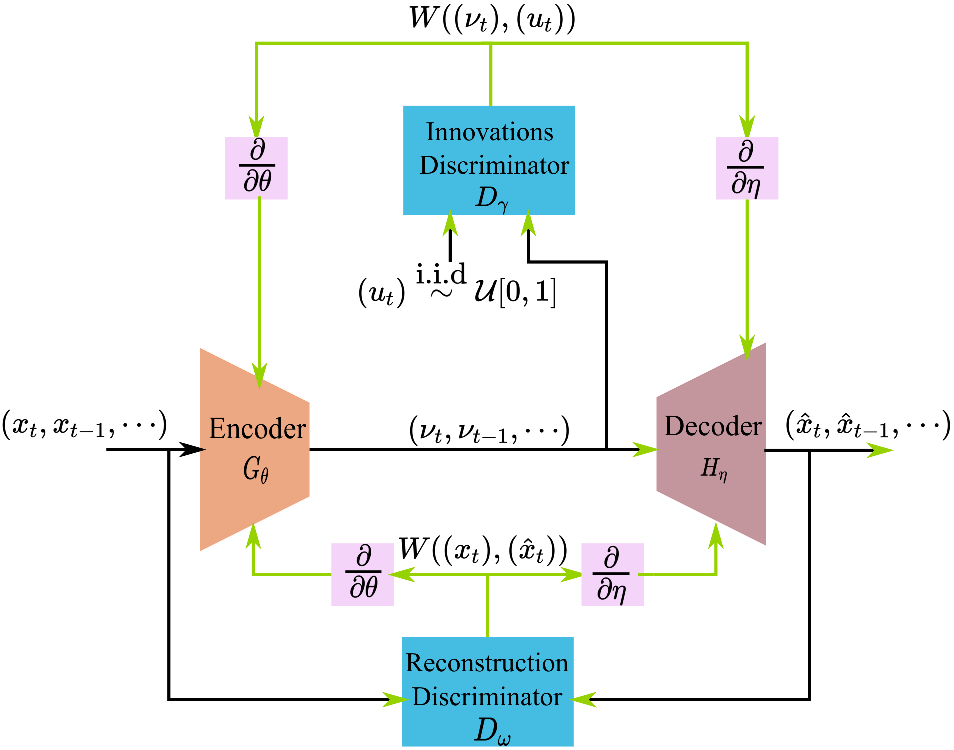}
    \caption{Training WIAE with discriminators.}
    \label{fig:training scheme}
    \vspace{-1.5em}
\end{figure}
This section focuses on the learning of WIAE that generates the weak innovation sequence as its latent process. Shown in Fig.~\ref{fig:training scheme} is the schematic, where both the encoder $G_\theta$ and decoder $H_\eta$ are convolutional neural networks parameterized by neural network coefficients $\theta$ and $\eta$, respectively.

The WIAE is trained to enforce the weak innovation conditions for its latent process using gradients from two discriminators: the {\em innovation discriminator $D_\gamma$} to force $(\nu_t)$ to be i.i.d. uniform and the {\em reconstruction discriminator $D_\omega$} to match the input-output distributions. We present the objective and an algorithm for WIAE learning in the following paragraphs.

We now define the objective and an algorithm for WIAE learning to generate weak innovation sequence $(V_t)$ as its latent process.
Both discriminators $D_\gamma$ and $D_\omega$ in Fig.~\ref{fig:training scheme} compare two random processes to produce gradient updates: $D_\gamma$ provides gradient signals for encoder $G_\theta$ to enforce the i.i.d. uniform condition on its output $(\nu_t)$, and $D_\omega$ provides gradient signals for the encoder-decoder pair $(G_\theta, H_\eta)$ to produce the input-output matching distributions for WIAE. It is then natural to define the training objective function by a weighted sum of training losses observed by the two discriminators.   

We use Wasserstein distance $W(f_1,f_2)$ to measure the distance between two distributions $f_1$ and $f_2$. In particular, we adopt the learning paradigm of Wasserstein GAN (WGAN) \cite{Arjovsky17} to train deep neural network discriminators $D_\gamma$ and $D_\omega$.
By the Kantorovich-Rubinstein duality, Wasserstein distance between two random variables $X\sim f_1$ and $Y\sim f_2$ can be written as the maximum expected difference between $D(X)$ and $D(Y)$ under a 1-Lipschitz function $D$:
\begin{align}
    W(f_1,f_2) = \max_{D: 1-Lipschitz}\left(\mbbE_{X\sim f_1}[D(X)]-\mbbE_{Y\sim f_2}[{D}(Y)]\right).
    \label{eq:wass}
\end{align}
WGAN employs a $\omega$-parameterized neural network for the 1-Lipschitz function $D_\omega$, with $\omega$ optimized according to \eqref{eq:wass}.

Specializing to WIAE learning in Fig.~\ref{fig:training scheme}, the two discriminators in WIAE are deep neural networks $D_\gamma$ and $D_\omega$ minimized jointly under the following risk measure:
\begin{multline}
     L((X_t),\theta,\eta):= \max_{\gamma,\omega}\big(\mbbE[D_{\gamma}(U_t) - \mbbE[D_\gamma(V_{t})]] \\
     + \lambda(\mbbE[D_{\omega}(X_{t})]
      -\mbbE[D_{\omega}(\hat{X}_{t})])\big).
  \label{eq:loss}
\end{multline}
$\lambda$ is a real number that scales the two Wasserstein distances.
The two parts of the loss function regularize the innovation according to \eqref{Eq:iid} and \eqref{Eq:recons}.
Minimizing Eq.~\eqref{eq:loss} with respect to $\theta$ and $\eta$ is thus equivalent to enforcing $(V_t)$ being i.i.d uniform, and $(\hat{X}_t)$ having the same distribution as $(X_t)$. 

%% file: simulation_final.tex
This section presents an application of GPF-WI to real-world time series forecasting problems.
We conducted numerical experiments on a benchmark M4 dataset and two real-time wholesale electricity price datasets for their high volatility.
The detailed information about electricity price datasets can be found in Sec.~\ref{subsec:electricity price}.
Among the M4 datasets, we chose the electricity consumption one (hereafter referred to as Electricity) consisting of hourly load measurements from $370$ clients.
\renewcommand{\arraystretch}{0.4}
\begin{table*}[t]
\fontsize{7.5}{1}
  \caption{Numerical Results. The numbers in the parentheses after each dataset name indicates the prediction step. The number in parentheses after each value of the metrics indicates the ranking of the algorithm.\\[0.5em]
  }
  \label{tb:Result-1}
  \centering
  \begin{tabular}{llllllllllll}
    \toprule
    \scriptsize Metrics &\scriptsize Method     &\scriptsize ISONE ($75$ min)    &\scriptsize ISONE ($2$ hour) &\scriptsize NYISO ($75$ min) &\scriptsize NYISO ($2$ hour) &\scriptsize Electricity ($5$ hour) &\scriptsize Electricity ($10$ hour)\\
    \midrule
       \scriptsize \multirow{6}{*}{\textbf{NMSE}} & \scriptsize GPF-WI &$\mathbf{0.0857}$ $\mathbf{(1)}$  & $\mathbf{0.0868}$ $\mathbf{(1)}$ &$\mathbf{0.0876}$ $\mathbf{(1)}$ &$\mathbf{0.0887}$ $\mathbf{(1)}$  &$0.4682$ $(3)$ &$0.5165$ $(3)$  \\
    &\scriptsize DeepAR    &$0.1064$ $(2)$ &$0.2852$ $(6)$&$0.0952$ $(2)$ &$0.1632$ $(3)$ &$0.5290$ $(4)$  &$0.6272$ $(4)$   \\
    &\scriptsize NPTS     &$0.1216$ $(3)$ &$0.1250$ $(2)$ &$0.1840$ $(5)$ &$0.2031$ $(5)$ & $1.3862$ $(6)$ &$1.3867$ $(6)$   \\
    &\scriptsize Pyraformer &$0.1247$ $(6)$ &$0.1779$ $(4)$ &$0.4031$ $(6)$ &$0.4229$ $(6)$ &$\mathbf{0.1422}$ $\mathbf{(1)}$ &$\mathbf{0.1735}$ $\mathbf{(1)}$\\
    &\scriptsize TLAE &$0.1219$ $(4)$ &$0.1837$ $(5)$&$0.1116$ $(3)$ &$0.1492$ $(2)$ &$0.1881$ $(2)$ &$0.2304$ $(2)$\\
    &\scriptsize Wavenet &$0.1245$ $(5)$ &$0.1643$ $(3)$ &$0.1128$ $(4)$ &$0.1809$ $(4)$ &$1.3367$ $(5)$ &$1.3575$ $(5)$\\
    &\scriptsize SNARX &$0.8632$ $(7)$ &$0.9658$  $(7)$&$0.9586$ $(7)$ &$1.8063$ $(7)$ &$1.7193$ $(7)$ &$1.8372$ $(7)$\\
    \midrule
    \scriptsize \multirow{6}{*}{\textbf{NMAE}}
    &\scriptsize GPF-WI &$\mathbf{0.2327}$ $\mathbf{(1)}$  & $\mathbf{0.2330}$ $\mathbf{(1)}$ &$0.2112$ $(3)$ &$\mathbf{0.2104}$ $\mathbf{(1)}$  &$\mathbf{0.2296}$ $\mathbf{(1)}$ &$\mathbf{0.3487}$ $\mathbf{(1)}$  \\
    &\scriptsize DeepAR    &$0.2969$ $(5)$&$0.4458$ $(6)$ &$0.1799$ $(2)$ &$0.2451$ $(4)$ &$0.3360$ $(3)$ &$0.3559$ $(2)$  \\
    &\scriptsize NPTS     &$0.2896$ $(4)$ &$0.2958$ $(2)$ &$0.2260$ $(4)$ &$0.2138$ $(2)$ & $0.4625$ $(5)$ &$0.4627$ $(5)$  \\
    &\scriptsize Pyraformer &$0.2570$ $(3)$ &$0.3518$ $(4)$ &$0.4892$ $(6)$ &$0.5234$ $(6)$ &$0.3396$ $(4)$ &$0.3794$ $(4)$\\
    &\scriptsize TLAE  &$0.2565$ $(2)$ &$0.3536$ $(5)$&$0.2258$ $(5)$ &$0.2590$ $(5)$ &$0.3041$ $(2)$ &$0.3597$ $(3)$\\
    &\scriptsize Wavenet  &$0.3154$ $(6)$ &$0.3310$ $(3)$&$\mathbf{0.1770}$ $\mathbf{(1)}$ &$0.2330$ $(3)$&$0.6215$ $(6)$ &$0.6216$ $(6)$\\
    &\scriptsize SNARX  &$0.8317$ $(7)$ &$0.9611$ $(7)$&$0.9299$ $(7)$ &$0.9967$ $(7)$ &$0.7243$ $(7)$ &$0.7624$ $(7)$\\
    \midrule
    \scriptsize \multirow{6}{*}{\textbf{sMAPE}}
    &\scriptsize GPF-WI &$0.2421$ $(2)$ & $\mathbf{0.2419}$ $\mathbf{(1)}$ &$0.2106$ $(4)$&$\mathbf{0.2091}$ $\mathbf{(1)}$  &$\mathbf{0.3266}$ $\mathbf{(1)}$ &$\mathbf{0.4034}$ $\mathbf{(1)}$ \\
    &\scriptsize DeepAR    &$\mathbf{0.2372}$ $\mathbf{(1)}$ &$0.7076$ $(5)$ &$\mathbf{0.1709}$ $\mathbf{(1)}$ &$0.2236$ $(2)$ &$0.4917$ $(3)$ &$0.5004$ $(3)$    \\
    &\scriptsize NPTS     &$0.8601$ $(7)$ &$0.8854$ $(6)$ &$0.7395$ $(6)$ &$0.7506$ $(6)$ &$0.6051$ $(4)$ &$0.6053$ $(4)$  \\
    &\scriptsize Pyraformer  &$0.2785$ $(4)$ &$0.2896$ $(2)$ &$0.3672$ $(5)$&$0.4095$ $(5)$ &$0.6385$ $(5)$ &$0.6546$ $(5)$ \\
    &\scriptsize TLAE   &$0.2440$ $(3)$ &$0.3137$ $(3)$ &$0.1974$ $(2)$&$0.2549$ $(4)$  &$0.3974$ $(2)$ &$0.4791$ $(2)$\\
    &\scriptsize Wavenet &$0.3057$ $(5)$ &$0.3375$ $(4)$ &$0.2090$ $(3)$ &$0.2475$ $(3)$ &$1.9877$ $(7)$ &$1.9858$ $(7)$\\
    &\scriptsize SNARX  &$0.5035$ $(6)$ &$1.3589$ $(7)$ &$1.2661$ $(7)$ &$1.8736$ $(7)$ &$1.0783$ $(6)$  &$1.1125$ $(6)$\\
    \midrule
    \scriptsize \multirow{6}{*}{\textbf{CRPS}}
    &\scriptsize GPF-WI &$\mathbf{0.4024}$ $\mathbf{(1)}$  & $0.4215$ $(2)$ &$\mathbf{0.0737}$  $\mathbf{(1)}$&$\mathbf{0.1038}$ $\mathbf{(1)}$ &$\mathbf{0.1451}$ $\mathbf{(1)}$ &$\mathbf{0.1944}$ $\mathbf{(1)}$  \\
    &\scriptsize DeepAR    &$0.4355$ $(4)$ &$0.4984$ $(4)$ &$0.0901$ $(2)$ &$0.1065$ $(2)$ &$1.8801$ $(3)$ &$0.1946$ $(2)$   \\
    &\scriptsize NPTS     &$0.4154$ $(2)$ &$\mathbf{0.4192}$ $\mathbf{(1)}$ &$0.1098$ $(3)$ &$0.1066$ $(3)$ & $0.1703$ $(2)$ &$0.2014$ $(3)$   \\
    &\scriptsize TLAE    &$0.4213$ $(3)$ &$0.4544$ $(3)$ &$0.7031$ $(5)$&$0.6912$ $(5)$ &$0.6776$ $(6)$ &$0.7662$ $(6)$\\
    &\scriptsize Wavenet &$0.4870$ $(5)$ &$0.6832$ $(5)$ &$0.2030$ $(4)$ &$0.2198$ $(4)$ &$0.3466$ $(4)$ &$0.3474$ $(4)$\\
    &\scriptsize SNARX  &$1.4593$ $(6)$ &$1.5629$ $(6)$ &$1.3264$ $(6)$ &$1.7342$ $(6)$ &$0.5263$ $(5)$ &$0.6271$ $(5)$\\
    \bottomrule
  \end{tabular}
\end{table*}

We compared our methods with $6$ other state-of-the-art time series forecasting techniques: DeepAR \cite{salinas_deepar_2019}, Nonparametric Time Series Forecaster (NPTS) \cite{alexandrov_gluonts_2019}, Pyraformer \cite{liu_pyraformer_2022}, TLAE \cite{nguyen_temporal_2021}, Wavenet \cite{oord_wavenet_2016}, and SNARX \cite{weron_forecasting_2008}.
Both probabilistic forecasting methods and point-forecast methods were included.
DeepAR is an auto-regressive RNN time series model that estimates the parameters of parametric distributions.
NPTS is a probabilistic forecasting technique that resembles naive forecasters.
It randomly samples a past time index following a categorical probability distribution over time indices.
TLAE is a nonparametric probabilistic forecasting technique that utilizes autoencoder architecture to learn the underlying Gaussian latent process, and uses it as the estimator.
Wavenet is a parametric probabilistic forecasting technique that is based on dilated causal convolutions.
Pyraformer is a point estimation technique that adopts multi-resolution attention modules, and is trained by minimizing mean squared error.
SNARX is a semiparametric AR model that utilizes kernel density function to estimate the distribution of noise, which has superior empirical performance on electricity pricing datasets according to \cite{weron_electricity_2014}.

To comprehensively compare the strengths and weaknesses of different forecasting techniques, we adopted multiple evaluation metrics: normalized mean square error (NMSE), normalized mean absolute error (NMAE), symmetric mean absolute percentage error (sMAPE), and continuous ranked probability score (CRPS).
For probabilistic forecasting methods, we used sample mean as the point estimator when calculating square errors, and sample median for absolute errors.
In total, $1000$ trajectories were sampled for probabilistic methods.
The commonly used mean absolute percentage error was not adopted, since in electricity price datasets the absolute value of the prices can be very close to $0$, which nullifies the effectiveness of the metrics.
Since electricity price datasets exhibit high variability, we excluded the outliers, which are defined as samples that are three standard deviations away from the sample mean, when calculating all metrics, for all methods.
\subsection{Electricity Prices and Datasets}
\label{subsec:electricity price}
\begin{figure}[t]
    \vspace{-2em}
    \centering
    \includegraphics[scale=0.4]{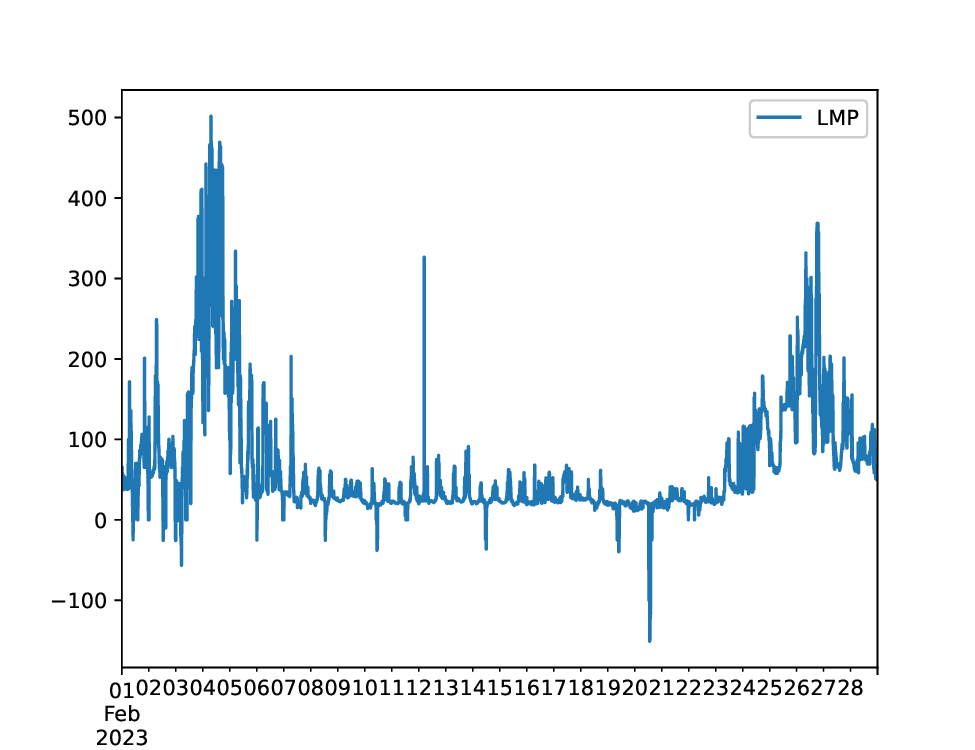}
    \caption{Trajectory of ISONE real-time electricity price.}
    \label{fig:price_traj}
    \vspace{-1.5em}
\end{figure}

Wholesale electricity prices in the U.S. are computed from an optimization over the power generation and demand levels based on offers from producers and bids from power generators, subject to generation limits and power flow constraints.
The price of electricity at a particular location is the Lagrange multiplier associated with the power balance constraint at that location, which is very sensitive to the line congestion pattern of the power grid.  When a line changes from uncongested to congested due to even a tiny change in power flow, the price of electricity incurs a sudden change.  Such discontinuities make electricity prices highly volatile.  See Fig.~\ref{fig:price_traj} for the price trajectory of a typical day.
The wholesale market price model can be found in \cite{Ji&Thomas&Tong:17TPS,Ji&Deng&Tong:21Bkchap}, where the authors also considered a  very different price forecasting problem where the system operator has access to the network operating conditions and offers and bids from market participants.
In contrast, the price forecasting problem studied here is time series forecasting without assuming how the time series is generated.
We chose two publicly available real-time electricity price datasets from two independent system operators (ISO) to demonstrate the ability of forecasting time series with higher sampling frequency and larger variability.
We named the datasets with the names of the ISOs: New York ISO (NYISO) and ISO New England (ISONE).
NYISO and ISONE consist of 5-minute real time electricity prices.
We took one month (February of 2023) of electricity prices for NYISO and ISONE.

We conducted 75-min and 2-hour ahead prediction for 5-min real-time electricity price (ISONE \& NYISO), and 5-hour and 10-hour ahead prediction for hourly sampled electricity consumption.
The WIAE is trained with Adam optimizer.

\subsection{Results}
The simulation result is shown in Table.~\ref{tb:Result-1}.
It can be seen that GPF-WI had better performance for most of the metrics.
Pyraformer, as a point estimation technique optimized by MSE, performed worse under metrics using absolute errors.
We also observed that the auto-regressive probabilistic forecasting methods (DeepAR, NPTS, Wavenet,SNARX) have the tendency to be affected by past observation, which leads to better performance when the time series is smooth with few fluctuations.
Due to the fact that electricity prices being a Lagrangian multiplier that doesn't exhibit continuity as other measurements taken from a physical system, the auto-regressive methods predicted shifted peaks, which contributed to large errors under more volatile datasets.
On the other hand, the generative probabilistic forecasting methods (GPF-WI, TLAE), where the prediction is conditioned on latent processes, suffered less from the volatility problem.
For TLAE, the latent process being a correlated Gaussian process posed great difficulty to the Monte Carlo sampling of the latent process, essential to probabilistic forecasting.
This was exhibited by the instability of probabilistic forecasts, which varies significantly with between trajectories. 

%% file: appendix.tex
\section{Proof of Theorem \ref{thm:converge}}
For any fixed $n$, we define the following notation:
\begin{align*}
V_{t,m}^{*(n)} &= G_{\theta_m^*}(X_t,\cdots,X_{t-m+1}),\\
\hat{X}_{t,m}^{*(n)} &= H_{\eta_m^*}(V_{t,m}^{(n)},\cdots,V^{(n)}_{t-m+1,m}),\\
\tilde{V}_{t,m}^{(n)} &= G_{\tilde{\theta}_m}(X_t,\cdots,X_{t-m+1}).
\end{align*}
We denote the set of weights that obtain optimality under Eq.~\eqref{eq:loss} by $(\theta_m^*,\eta_m^*,\gamma_m^*,\omega_m^*)$. 
We make the clear distinction between the true weak innovations $\{V_t\}$ and estimated weak innovations with $m,n$-dimensional Weak Innovations Auto-encoder (WIAE), which is denoted by $\{V_{t,m}^{(n)}\}$.

Denote the $n$-dimensional vector $[V_{t,m}^{*(n)},\cdots,V_{t-n+1,m}^{*(n)}]$ by $\boldsymbol{V}_{t,m}^{*(n)}$, with the letter changed to bold face.
${\boldsymbol U}_{t,m},\hat{\boldsymbol X}_{t,m}^{(n)}$ and $\boldsymbol X^{(n)}_{t,m}$ are defined similarly. $\{\tilde{V}_t\}$ is the sequence generated by $G_{\tilde{\theta}_m}$. All the random variables generated by $(\tilde{\theta}_m,\tilde{\eta}_m)$ are defined in the similar pattern.

{\em Proof:} 
\ben
\item Want to show that
$L(\tilde{\theta}_m,\tilde{\eta}_m)\rightarrow 0$ as $m\rightarrow \infty$.

By assumption A2, $G_{\tilde{\theta}_m}\rightarrow G$ uniformly, thus $\lVert \tilde{\boldsymbol V}_{t,m}^{(n)}-\boldsymbol V_t^{(n)}\rVert<n\epsilon$ for $\forall \epsilon>0$. 
Thus $\tilde{\boldsymbol V}_{t,m}^{(n)}\stackrel{d}{\Rightarrow}\boldsymbol V_t$. 
Similarly, we have $[H_{\tilde{\eta}_m}(V_{t},\cdots,V_{t-m+1}),\cdots,H_{\tilde{\eta}_m}(V_{t-n+1},\cdots,V_{t-n-m+2})]\stackrel{d}{\Rightarrow} [X_t,\cdots,X_{t-n+1}]$.

Since $H$ is continuous and $H_{\tilde{\eta}_m}$ converge uniformly to $H$, $H_{\tilde{\eta}_m}$  is continuous. Thus by continuous mapping theorem, $H_{\tilde{\eta}_m}(\tilde{V}_{t,m}^{(n)},\cdots,\tilde{V}_{t-m+1,m}^{(n)})\stackrel{d}{\Rightarrow}H_{\tilde{\eta}_m}(V_{t},\cdots,\nu_{t-m+1})$. Thus, 
\[[H_{\tilde{\eta}_m}(\tilde{V}_{t,m}^{(n)},\cdots,\tilde{V}_{t-m+1,m}^{(n)}),\cdots,H_{\tilde{\eta}_m}(\tilde{V}_{t-n+1,m}^{(n)},\cdots,\tilde{V}^{(n)}_{t-n-m+2,m})]\stackrel{d}{\Rightarrow} [X_t,\cdots,X_{t-n+1}]\] 
Hence $L(\tilde{\theta}_m,\tilde{\eta}_m)\rightarrow 0$.

Therefore, $L(\theta_m^*,\eta_m^*):=\min_{\theta,\eta}L(\theta,\eta) \leq L(\tilde{\theta}_m,\tilde{\eta}_m)\rightarrow 0$.

\item Since $L^{(n)}(\theta_m^*,\eta_m^*)\rightarrow 0$ as $m\rightarrow \infty$, by Eq.~\eqref{eq:loss}, $\boldsymbol V_{t,m}^{(n)} \stackrel{d}{\Rightarrow} \boldsymbol U_{t,m}^{(n)}$, $\hat{\boldsymbol X}_{t,m}^{(n)}\stackrel{d}{\Rightarrow}\boldsymbol X^{(n)}_{t,m}$ follow directly by the equivalence of convergence in Wasserstein distance and convergence in distribution \citep{Villani09:Book}.
\een

\section{Proof of Lemma \ref{thm:sufficiency}}
{\em Proof:} By the definition of weak innovations, $\{X_t\}\stackrel{d}{=}\{\hat{X}_t\}$. Hence we know that the conditional cumulative density function (CDF) of $\{\hat{x}_t\}$ satisfies
\begin{multline}
    \mathbb{P}[\hat{X}_{t+T}\leq x|\hat{X}_{t}=x_{t},\hat{X}_{t-1}=a_{t-1},\cdots]=\mathbb{P}[X_{t+T}\leq x|X_{t}=x_{t},X_{t-1}=x_{t-1},\cdots]=F(x|\xbf_t).
\end{multline}
Since the infinite dimensional decoder function $H$ is injective,
\[\mathbb{P}[\hat{X}_{t+T}\leq x|\xbf_t] = \mathbb{P}[\hat{X}_t\leq x|V_{t-1}=\nu_{t-1},V_{t-2}=\nu_{t-2},\cdots],\]
where $\nu_{s} = H(x_s,x_{s-1},\cdots)$.
Hence lemma \ref{thm:sufficiency} holds.

\section{Performance Evaluation of WIAE on Representation Extraction}
\label{subsec:rep-learning}
In this section, we conducted property tests to test the capability of our method to produce weak innovations representation.
We performed numerical experiments on three different synthetic datasets named as LAR, MAR and MC. 
The specific formulation of the synthetic datasets is shown in Table.~\ref{tab:Synthetic dataset}.
The three synthetic cases were designed to cover three different existence scenarios of the innovations representation.
The Linear Auto Regression (LAR) case has known innovations sequence \citep{WangTong:21JMLR} that can be extracted through linear function pair.
Consequently, the weak innovations sequence must exist for the LAR case.
On the other hand, the MA case satisfies the existence condition for weak innovations sequence.
Since the MA case is of minimum phase, whether it has innovations representation remains agnostic.   
The two-state Markov Chain case (MC) with transition probability specified in Table.~\ref{tab:Synthetic dataset}, only has weak innovations as shown by \cite{Rosenblatt:59}.

\begin{table}[t]
\caption{Test Synthetic Datasets. $u_t\stackrel{\tiny\rm i.i.d}{\sim}\mathcal{U}[-1,1]$.}
\label{tab:Synthetic dataset}
\begin{center}
\begin{small}
\begin{sc}
\begin{tabular}{ll}
\toprule
Dataset & Model \\
\midrule
Moving Average (MA)    &$x_t=u_t+2.5u_{t-1}$  \\
Linear Autoregressive (LAR) &$x_t=0.5 x_{t-1}+u_t$ \\
Two-State Markov Chain (MC)    &$P=  \begin{bmatrix} 0.6 & 0.4\\0.4 &0.6\end{bmatrix}$\\
\bottomrule
\end{tabular}
\end{sc}
\end{small}
\end{center}
\vspace{-2em}
\end{table}

We first ran property tests to evaluate the performance in terms of extracting weak innovations representation from original samples. 
Here we adopted the hypothesis testing formulation known as the Runs Up and Down test \citep{Gibbons:03Book}. 
Its null hypothesis assumes the sequence being i.i.d, and the alternative hypothesis the opposite. 
The test was based on collecting the number of consecutively increasing or decreasing subsequences, and then calculate the p-value of the test based on its asymptotic distribution.  According to \citet{Gibbons:03Book}, the runs up and down test had empirically the best performance. 

We compared with $5$ representation learning benchmarks: fAnoGAN \citep{Schlegl&Seebock:19}, Anica \citep{Brakel&Bengio:17}, IAE \citep{WangTong:21JMLR}, Pyraformer \citep{liu_pyraformer_2022}, TLAE \citep{nguyen_temporal_2021}, on both synthetic datasets (LAR, MA, MC) and real-world datsets (ISONE, NYISO,{\it SP500}).
LAR, MA, MC stands for Linear Autoregression, Moving Average and Markov Chain models, respectively.
For the LAR case, \cite{Wu:05PNAS} showed that the original form of innovations (Eq.~(\ref{eq:s-encoder}-\ref{eq:s-iid})) exist, while for the MA case that is non-minimum phase, the existence remains unknown.
For the MC case, \cite{Rosenblatt:59} proved that the original form of the innovations representation doesn't exist, though the weak innovations do.
More details of benchmarks and datasets can be found in the appendix.
\renewcommand{\arraystretch}{0.5}
\begin{table*}[t]
\fontsize{6}{0.1}
    \centering
    \caption{P-value from Runs Test, Wasserstein distance (WassDist-rep) bewteen representations and uniform {\it i.i.d} sequence, and Wasserstein distance between the original sequence and reconstruction (WassDist-recons).}
    \begin{tabular}{ccccccc}
    \toprule
       {\scriptsize Metrics} & {\scriptsize Datasets} & {\scriptsize WIAE} & \scriptsize IAE &\scriptsize fAnoGAN &\scriptsize Anica &\scriptsize TLAE  \\
        \midrule
         \multirow{6}{*}{\scriptsize P-value} 
         &\scriptsize LAR   &$\mathbf{0.9513}$ &$0.8837$ &$0.0176$   &$<0.001$   &$0.4902$\\
         &\scriptsize MA    &$0.8338$ &$\mathbf{0.9492}$ &$0.2248$   &$<0.001$   &$0.5498$\\
         & \scriptsize MC    &$\mathbf{0.7651}$ &$<0.001$ &$0.2080$   &$<0.001$   &$<0.001$\\
         &\scriptsize ISONE &$\mathbf{0.8897}$ &$<0.001$ &$<0.001$   &$0.2610$   &$0.6731$\\
         & \scriptsize NYISO &$\mathbf{0.7943}$ &$<0.001$ &$<0.001$   &$<0.001$   &$0.2269$\\
         &\scriptsize SP500 &$\mathbf{0.8764}$ &$<0.001$ &$0.4977$   &$0.2655$   &$0.8183$\\
         \midrule
         \multirow{6}{*}{\shortstack{\scriptsize WassDist-rep\\$\pm$ \scriptsize std}} 
         &\scriptsize LAR   &$\mathbf{0.3651\pm0.0033}$  &$0.5555\pm0.0551$  &$0.9289\pm0.0207$  &$0.8698\pm0.0186$  &$0.9525\pm0.1697$\\
         &\scriptsize MA    &$\mathbf{0.3563\pm0.0053}$  &$0.4839\pm0.0202$  &$1.0679\pm0.0339$  &$0.8518\pm0.0740$  &$0.7346\pm0.2502$\\
         &\scriptsize MC    &$0.3814\pm0.0053$  &$0.3984\pm0.0687$  &$1.3093\pm0.0276$  &$0.7112\pm0.0231$  &$\mathbf{0.0906\pm0.6693}$\\
         &\scriptsize ISONE &$\mathbf{0.0391\pm0.0243}$  &$0.3796\pm0.2080$  &$0.0959\pm0.0980$  &$1.0579\pm0.1328$  &$0.0814\pm0.0745$\\
         &\scriptsize NYISO &$\mathbf{0.0405\pm0.0306}$  &$0.3192\pm0.1186$  &$0.1102\pm0.0322$  &$1.4802\pm0.1069$  &$0.0678\pm0.0239$\\
         &\scriptsize SP500 &$\mathbf{0.1287\pm0.0036}$  &$0.3778\pm0.1904$  &$0.1341\pm0.0538$  &$1.1658\pm0.1398$  &$0.1755\pm0.1478$\\
         \midrule
         \multirow{6}{*}{\shortstack{\scriptsize WassDist-recons\\$\pm$ \scriptsize std}} 
         &\scriptsize LAR   &$1.7399\pm0.1464$  &$\mathbf{1.7119\pm0.1110}$  &$2.5150\pm1.0666$  &$4.0758\pm1.1990$  &$1.7625\pm0.1280$\\
         &\scriptsize MA    &$\mathbf{2.0214\pm0.1063}$  &$2.6697\pm0.5221$  &$10.9907\pm0.2981$ &$19.3772\pm0.8062$ &$4.6101\pm0.1878$\\
         &\scriptsize MC    &$\mathbf{0.1960\pm0.0338}$  &$2.8264\pm0.0996$  &$11.5251\pm1.7173$ &$12.4828\pm1.5741$ &$0.2639\pm0.1545$\\
         &\scriptsize ISONE &$\mathbf{0.3382\pm0.0695}$  &$0.8505\pm0.0032$  &$1.1341\pm0.05174$ &$2.0250\pm0.1790$  &$0.7436\pm0.1592$\\
         &\scriptsize NYISO &$0.1217\pm0.0029$  &$\mathbf{0.1111\pm0.0070}$  &$1.4021\pm0.07022$ &$2.3860\pm0.1307$  &$1.4098\pm0.3289$\\
         &\scriptsize SP500 &$\mathbf{0.1677\pm0.0551}$  &$1.1021\pm0.1904$  &$0.8563\pm0.2455$  &$1.8874\pm0.3518$  &$1.6977\pm0.4224$\\
         \bottomrule
    \end{tabular}
    \label{tab:property}
\end{table*}

The results of property tests are shown in Table.~\ref{tab:property}. 
Seen from the results, only WIAE had the capability of consistently producing i.i.d sequences that couldn't be rejected by the runs test for all cases, with small Wasserstein distances between representation and uniform i.i.d  sequence, and between the original series and reconstruction.
For all other techniques, at lease one case exists for which either the sequences they produced can be easily rejected by the runs test, or the sequences had high Wasserstein distances.
This indicated that the WIAE architecture design is necessary for the extraction of weak innovations representation, and that currently no such technique that fulfills the requirement apart from WIAE.
 
The Wasserstein distances between the representation sequence $\{\nu_t\}$ generated by each techniques and an {\it i.i.d} uniform sequence were also calculated to measure the independence of the representation. 
The Wasserstein distances of reconstruction are presented in Table.~\ref{tab:property} indicated by {\em WassDist-rep}. 
The Wasserstein distance were calculated through implementing a Wasserstein discriminator, with the same structure as all other neural networks. 
WIAE was able to achieve the best Wasserstein distance among all other techniques, with IAE the second best.

Similarly, we calculate the Wasserstein distance between reconstruction $\{\hat{x}_t\}$ and the original time series $\{x_t\}$.
For the LAR case, since the innovations sequence exists, IAE had the best performance. fAnoGAN was able to achieve reconstruction error close to WIAE for LAR case, yet its extracted components were easily rejected by runs test. 
For the MC case where only the weak innovations sequence exists, WIAE was the only technique able to produce significantly lower reconstruction loss (in terms of Wasserstein distance). Anica was proved to be less competitive in terms of decoded error.

\section{Definition of Metrics for Time Series Forecasting}
\label{sec:metrics}
In this section, we present the formulas of metrics used in the simulation section.
Let $\{x_t\}$ denote the original times series, $\{\hat{x}_t\}$ the prediction estimates, $N$ the size of datasets and $s$ the prediction step.
\begin{align}
    NMSE = \frac{\frac{1}{N}\sum_{t=1}^N(x_t-\hat{x}_t)^2}{\frac{1}{N}\sum_{t=1}^N x_t^2},\\
    NMeSE = med\left(\left(\frac{(x_t-\hat{x}_t)^2}{\frac{1}{N}\sum_{t=1}^N x_t^2}\right)_t\right),\\
    NMAE = \frac{\frac{1}{N}\sum_{t=1}^N|x_t-\hat{x}_t|}{\frac{1}{N}\sum_{t=1}^N|x_t|},\\
    NMeAE = med\left(\left(\frac{|x_t-\hat{x}_t|}{\frac{1}{N}\sum_{t=1}^N|x_t|}\right)_t\right),\\
    MASE = \frac{\frac{1}{N}\sum_{t=1}^N|x_t-\hat{x}_t|}{\frac{1}{N-s}\sum_{t=s+1}^N|x_t-x_{t-s}|},\\
    sMAPE = \frac{1}{N}\sum_{t=1}^N\frac{|x_t-\hat{x}_t|}{(|x_t|+|\hat{x}_t|)/2},
\end{align}
where $med(\cdot)$ denotes the operation of taking median of a given sequence.
The purpose of adopting multiple metrics is to comprehensively evaluate the level of performance for each method.
Especially for electricity price datasets with high variability, the methods with best median errors and best mean errors could be very different.
Mean errors, i.e., NMSE and NMAE, represent the overall performance, but are not robust to outliers.
On the other hand, median errors (NMeSE and NMeAE) are robust to outliers. 
MASE reflects the relative performance to the naive forecaster, which utilizes past samples as forecast.
Methods with MASE smaller than $1$ outperform the naive forecaster.
sMAPE is the symmetric counterpart of mean absolute percentage error (MAPE) that can be both upper bounded and lower bounded.
Since for electricity datasets, the actual values can be very close to $0$, thus nullifies the effectiveness of MAPE, we though that sMAPE is the better metric.

\section{Implementation Details of WIAE and Benchmarks}
DeepAR, NPTS and Wavenet are constructed from GluonTS \footnote{\url{https://ts.gluon.ai/stable/\#}} time series forecasting package.
We implemented WIAE, TLAE, IAE, Anica and fAnoGAN by ourselves. \footnote{WIAE can be found on the using the github link: \url{https://github.com/Lambelle/WIAE.git}.} for comparison. 
All the neural networks (encoder, decoder and discriminator) in the paper had three hidden layers, with the 100, 50, 25 neurons respectively. In the paper, $m=20, n=50$ was used for all cases. The encoder and decoder both used hyperbolic tangent activation. The first two layers of the discriminator adopted hyperbolic tangent activation, and the last one linear activation.

In training we use Adam optimizer with $\beta_1=0.9$, $\beta_2=0.999$. Batch size and epoch are set to be $60$ and $100$, respectively. The detailed hyperparameter of choice can be found in Table.~\ref{tb:Hyperparameters}. We note that under different version of packages and different cpus, the result might be different (even if we set random seed). The experimental results are obtained using dependencies specified in Github repository\footnote{\url{https://github.com/Lambelle/WIAE.git}}.
\begin{table}[t]
\caption{Hyper Parameters Setting for Each Case}
\label{tb:Hyperparameters}
\begin{center}
\begin{small}
\begin{sc}
\begin{tabular}{lcccl}
\toprule
\textbf{Test Case} & \textbf{Learning Rate} $\mathbf{\alpha}$ & \textbf{Gradient Penalty} $\mathbf{\lambda_1,\lambda_2}$ &\textbf{Weight for Reconstruction} ($\lambda$) \\
\midrule
MC   &$0.0001$ &$1.0,1.0$ &$1.0$  \\
AR1    &$0.0001$   &$1.0,1.6$ &$1.0$ \\
MA     &$0.0001$  &$1.0,1.6$ &$1.0$\\
SP500   &$1e-5$ &$1.0,1.3$  &$1.0$\\
NYISO   &$1e-5$    &$1.0,1.4$ &$1.0$\\
ISONE   &$1e-5$     &$1.0,1.0$ &$1.0$\\
PJM     &$1e-5$ &$1.0,1.0$ &$1.0$\\
electricity &$1e-5$ &$1.0,1.0$ &$1.0$\\
Traffic   &$1e-5$ &$1.0,1.2$ &$1.0$\\
\bottomrule
\end{tabular}
\end{sc}
\end{small}
\end{center}
\end{table}
\subsection{Convergence of WIAE Training}
\begin{figure}
    \centering
    \includegraphics[scale=0.5]{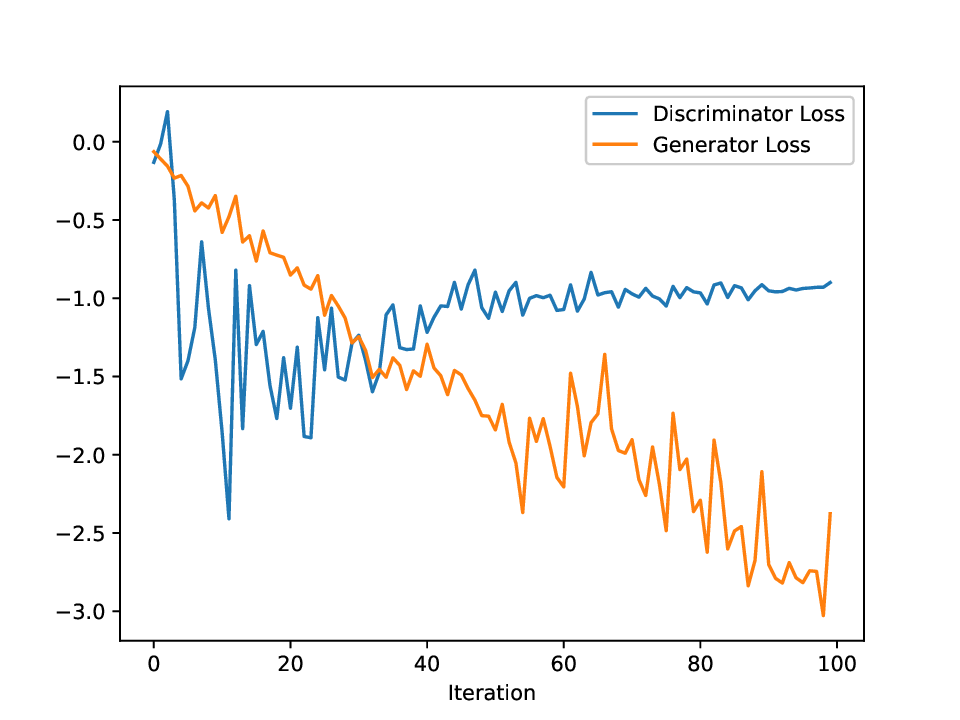}
    \caption{Training Losses vs Iterations}
    \label{fig:loss}
\end{figure}

In this section, we discuss about the GAN-based training algorithm of WIAE. 
Fig.~\ref{fig:loss} shows the loss variation with iteration for the training process on ISONE dataset.
Losses for other datasets exhibit similar pattern to some extent. 
We noticed that under some hyperparameters the convergence of training may no longer be guarateed, but overall the GAN-based training structure is sufficient for forecasting purpose.
This can be also illustrated by the standard deviation of experiments showed by the performance of WIAE on time series forecasting.

\section{Additional Time series Forecasting Results}
{\it Electricity} dataset consists of $15$-minute electricity consumption from $370$ households; {\it electricity} consists of hourly traffic sampled from $963$ car lanes; {\it SP500} contains daily opening and closing prices of S\&P500 stocks.
\renewcommand{\arraystretch}{0.5}
\begin{table}
  \caption{Numerical Results Continued. The numbers in the parenthese indicate the prediction step.}
  \label{tb:Result-2}
  \centering
  \begin{tabular}{llllllllll}
    \toprule
    Metrics &Method     & Electricity($5$)    & Electricity($10$) & Traffic($8$) &Traffic($20$) &SP500($2$) \\
    \midrule
       \multirow{6}{*}{\textbf{NMSE}} &WIAE &$0.5165$  & $0.5165$ &$0.1091$ &$0.1299$  &$0.0049$   \\
    &DeepAR    &$0.5290$ &$0.6272$ &$0.0309$ &$0.0797$ &$\mathbf{0.0002}$ \\
    &NPTS     &$1.3862$ &$1.3867$ &$0.3045$ &$0.3050$ & $0.0003$  \\
    &Pyraformer &$\mathbf{0.1422}$ &$\mathbf{0.1735}$ &$0.2722$ &$0.3744$ &$0.0050$\\
    &TLAE &$0.1881$ &$0.2304$ &$0.1097$ &$0.2790$ &$0.0013$\\
    &Wavenet &$1.3367$ &$1.3575$ &$\mathbf{0.0211}$ &$\mathbf{0.0241}$ &$0.0236$\\
    \midrule
    \multirow{6}{*}{\textbf{NMeSE}}
    &WIAE &$0.1416$  & $0.1626$ &$0.0237$ &$0.0578$ &$0.0017$    \\
    &DeepAR    &$0.1280$ &$0.1500$ &$\mathbf{0.0046}$ &$0.0200$ &$0.0001$ \\
    &NPTS     &$0.9135$ &$0.9136$ &$0.0893$ &$0.0894$ & $\mathbf{0.00003}$   \\
    &Pyraformer &$0.0139$ &$0.0181$ &$0.1194$ &$0.1830$ &$0.0047$ \\
    &TLAE &$\mathbf{0.0135}$ &$\mathbf{0.0151}$ &$0.0261$ &$0.1508$ &$0.0007$\\
    &Wavenet &$1.4559$ &$1.4895$ &$0.0211$ &$\mathbf{0.0036}$ &$0.0003$\\
    \midrule
    \multirow{6}{*}{\textbf{NMAE}}&WIAE &$0.3296$  & $\mathbf{0.3487}$ &$0.2350$ &$0.3012$  &$0.0538$ \\
    &DeepAR    &$0.3360$ &$0.3559$ &$0.1144$ &$0.2113$ &$0.0108$ \\
    &NPTS     &$0.4625$ &$0.4627$ &$0.4060$ &$0.4065$ & $0.0100$   \\
    &Pyraformer &$0.3396$ &$0.3794$ &$0.4544$ &$0.5293$ &$0.4819$\\
    &TLAE  &$\mathbf{0.3041}$ &$0.3597$ &$0.2710$ &$0.4641$ &$0.0313$\\
    &Wavenet  &$0.6215$ &$0.6216$ &$\mathbf{0.0864}$ &$\mathbf{0.0940}$ &$0.0552$\\
    \midrule
    \multirow{6}{*}{\textbf{NMeAE}}
    &WIAE &$0.2619$  & $0.2461$ &$0.1602$ &$0.2607$  &$0.0417$   \\
    &DeepAR    &$0.2332$ &$0.2533$ &$0.0738$ &$0.1562$ &$0.0085$    \\
    &NPTS     &$0.6190$ &$0.6191$ &$0.3244$ &$0.3245$ &$\mathbf{0.0056}$\\
    &Pyraformer &$0.1755$ &$0.2003$ &$0.3754$ &$0.4663$ & $0.1926$   \\
    &TLAE  &$\mathbf{0.1571}$ &$\mathbf{0.1729}$ &$0.1793$ &$0.4314$ &$0.0273$ \\
    &Wavenet &$0.7084$ &$0.7085$ &$\mathbf{0.0596}$ &$\mathbf{0.0627}$ &$0.0178$ \\
    \midrule
    \multirow{6}{*}{\textbf{MASE}}
    &WIAE &$0.8309$  & $\mathbf{0.8565}$ &$0.7948$  &$0.9208$  &$6.4547$ \\
    &DeepAR    &$0.8721$ &$0.9717$ &$0.5884$ &$0.5712$ &$1.2085$   \\
    &NPTS     &$1.996$ &$1.1151$ &$1.8938$ &$1.0654$ & $\mathbf{1.1475}$   \\
    &Pyraformer  &$1.0897$ &$1.0723$ &$1.9028$ &$1.3451$ &$17.4409$\\
    &TLAE   &$\mathbf{0.8002}$ &$0.9865$ &$0.9999$ &$0.9999$ &$1.4090$  \\
    &Wavenet &$2.6178$ &$2.4475$ &$\mathbf{0.4913}$ &$\mathbf{0.2945}$ &$6.9792$\\
    \midrule
    \multirow{6}{*}{\textbf{sMAPE}}
    &WIAE &$0.5274$  & $0.5034$ &$0.2861$ &$0.3715$  &$0.0540$ \\
    &DeepAR    &$0.4917$ &$0.5004$ &$0.1365$ &$0.2574$ &$
    \mathbf{0.0106}$    \\
    &NPTS     &$0.6051$ &$0.6053$ &$0.4871$ &$0.4876$ & $0.0518$  \\
    &Pyraformer  &$0.6385$ &$0.6546$ &$0.5489$ &$0.6264$ &$0.2785$ \\
    &TLAE   &$\mathbf{0.3974}$ &$\mathbf{0.4791}$ &$0.3184$ &$0.4995$ &$0.0298$\\
    &Wavenet &$1.9877$ &$1.9858$ &$\mathbf{0.1069}$ &$\mathbf{0.1164}$ &$0.0480$\\
    \midrule
    \multirow{6}{*}{\textbf{CRPS}}
    &WIAE &$0.1451$  & $0.1944$ &$0.0695$ &$0.0737$  &$0.2819$  \\
    &DeepAR    &$\mathbf{0.0881}$ &$\mathbf{0.0946}$ &$0.0938$ &$0.0760$ &$\mathbf{0.2598}$    \\
    &NPTS     &$0.1703$ &$0.1703$ &$0.0210$ &$0.0210$ & $0.6938$  \\
    &TLAE   &$0.6776$ &$0.7662$ &$0.3326$ &$0.7696$ &$0.8105$\\
    &Wavenet &$0.3466$ &$0.3474$ &$\mathbf{0.0044}$ &$\mathbf{0.0046}$ &$0.3920$ \\
    \bottomrule
  \end{tabular}
\end{table}

Due to page limitation, we show additional experimental results in this section.
Table.~\ref{tb:Result-2} shows the results for PJM, {\it traffic} and {\it electricity} datasets.
{\it Traffic} and {\it electricity} datasets are rather smooth measurements collected from physical systems, where Wavenet, the auto-regressive model performed relatively well.
PJM dataset contains hourly sampled electricity prices that are hard to predict. 
For different metrics, different methods performs well.

\section{Empirical CDF of Errors}

\begin{figure}[t]
    \centering
    \includegraphics[scale=0.25]{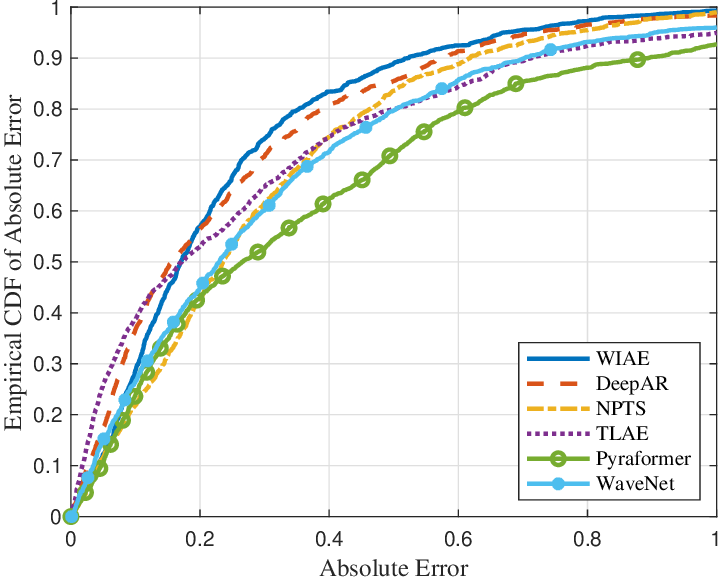}
    \includegraphics[scale=0.25]{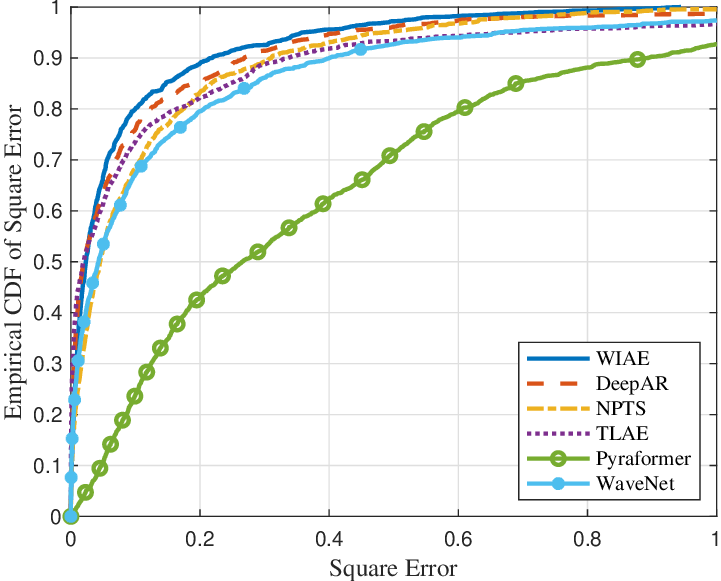}
    \includegraphics[scale=0.25]{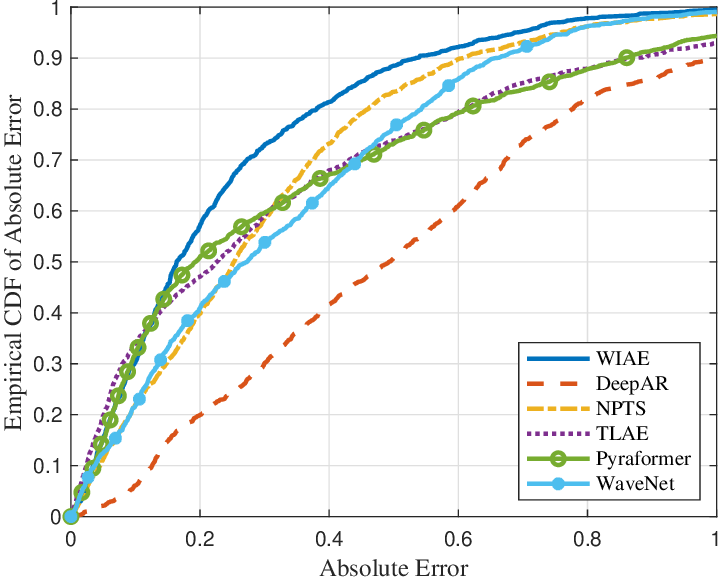}
    \includegraphics[scale=0.25]{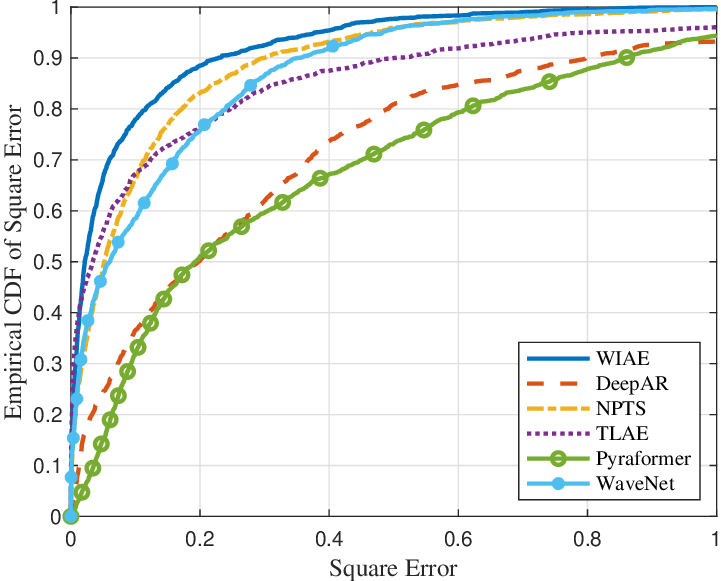}
    \caption{Errors for ISONE 15-step (two subplots to the left) and 24-step (two subplots to the right) prediction.}
    \label{fig:ecdf_isone}
\end{figure}

\begin{figure}[t]
    \centering
    \includegraphics[scale=0.25]{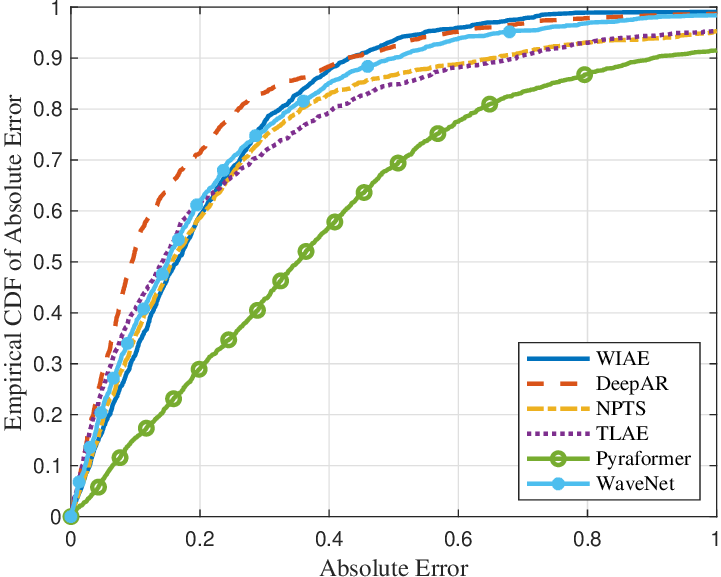}
    \includegraphics[scale=0.25]{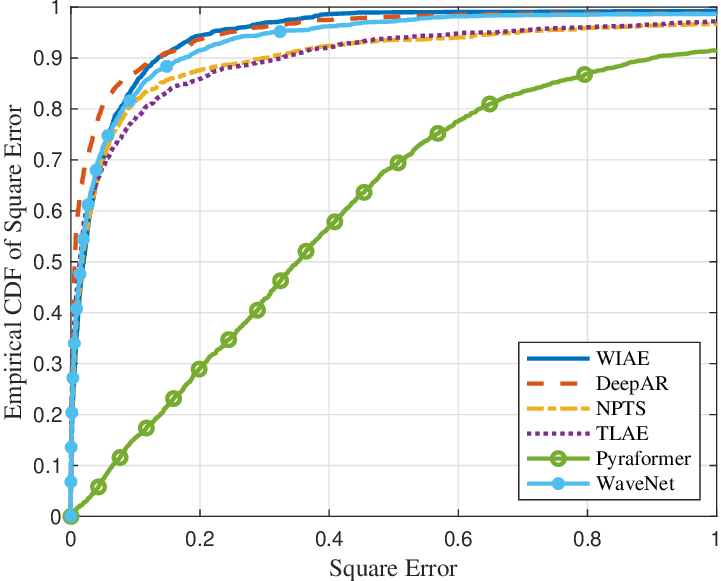}
    \includegraphics[scale=0.25]{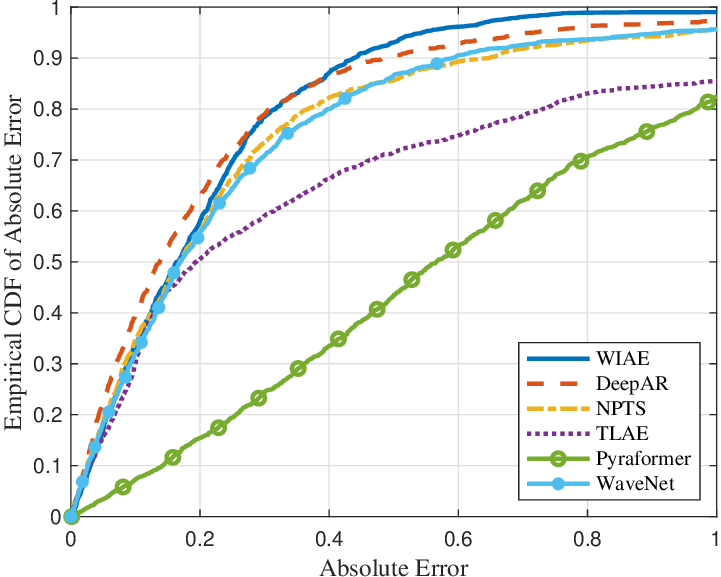}
    \includegraphics[scale=0.25]{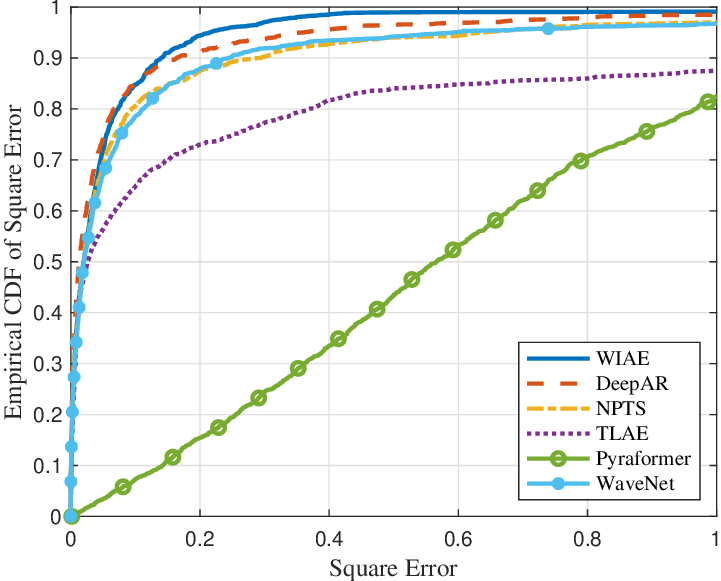}
    \caption{Errors for NYISO 15-step (two subplots to the left) and 24-step (two subplots to the right) prediction.}
    \label{fig:ecdf_nyiso}
\end{figure}

\begin{figure}[t]
    \centering
    \includegraphics[scale=0.25]{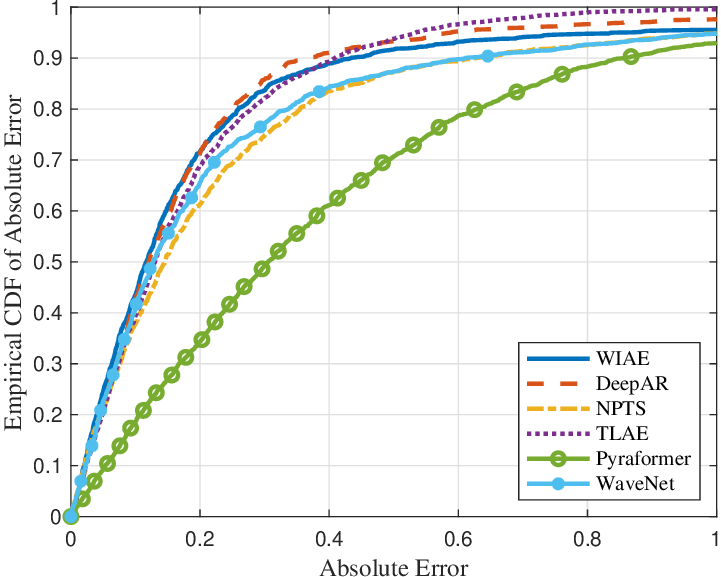}
    \includegraphics[scale=0.25]{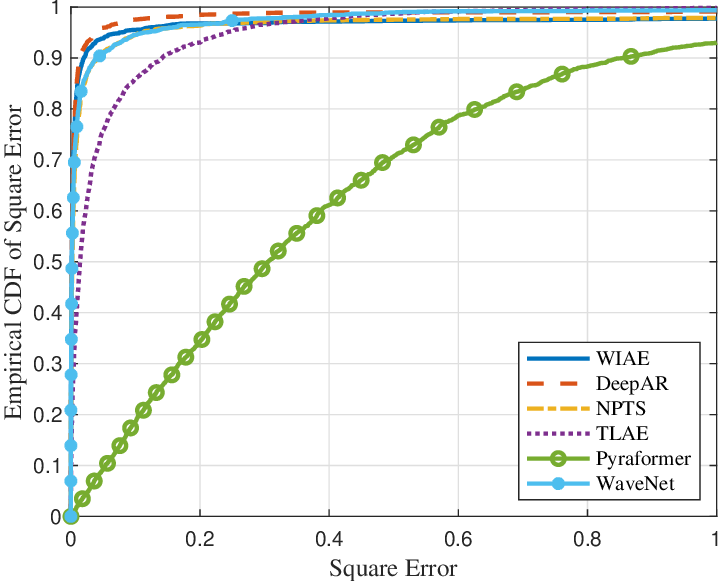}
    \includegraphics[scale=0.25]{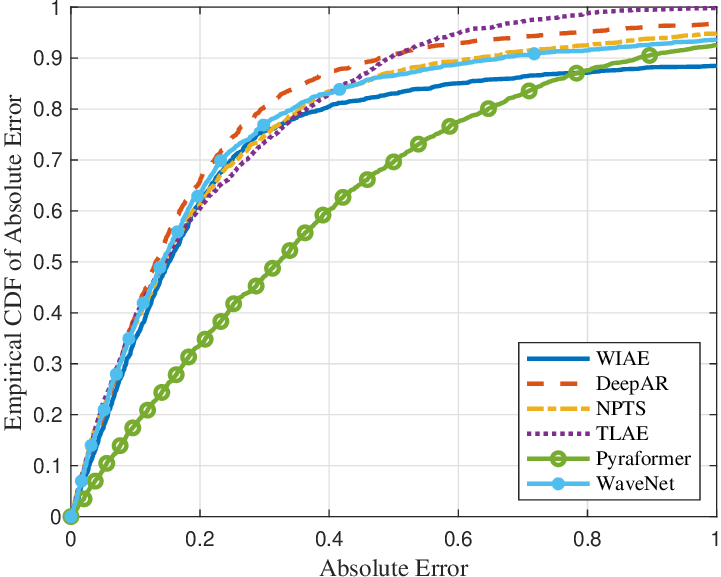}
    \includegraphics[scale=0.25]{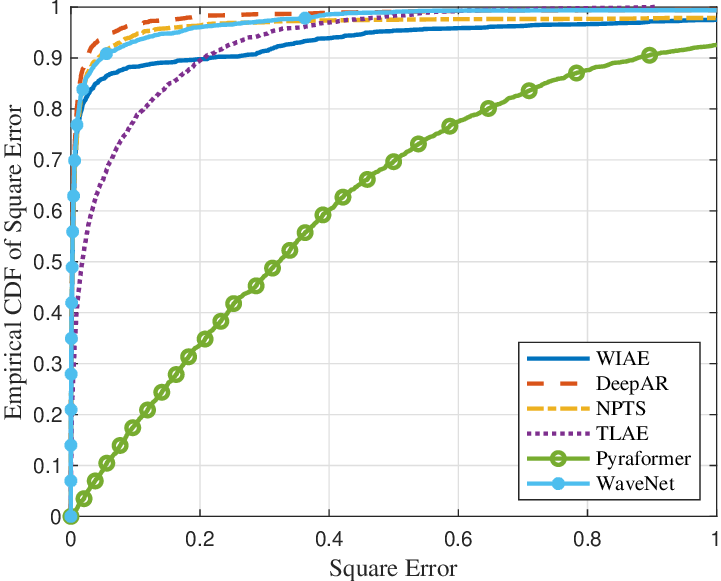}
    \caption{Errors for PJM 5-step (two subplots to the left) and 12-step (two subplots to the right) prediction.}
    \label{fig:ecdf_pjm}
\end{figure}

The metrics listed in Sec.~\ref{sec:metrics} are not able to comprehensively represent the performance of different methods, for mean and median are only two statistics that can be calculated from the error sequence of each method for different datasets. 
Especially for electricity price datasets with high variability, the distribution of error is quite skewed, which means mean and median errors can be quite different.
To provide a more comprehensive view of how different methods performed on electricity datasets, we plotted the empirical cumulative function (eCDF) of errors for each datasets.
The eCDF of an $N$-point error array (denoted by $F_N(\cdot)$) can be computed given:
\[F_N(t) = \frac{1}{N}\sum_{i=1}^N\mathbbm{1}_{e_i\leq t},\]
where $(e_i)$ is the error sequence, and $\mathbbm{1}$ the indicator function.
Hence, if the eCDF of a method is goes to $1$ when $t$ is small, then it means that the majority of the errors are small.

Fig.~\ref{fig:ecdf_isone}-\ref{fig:ecdf_pjm} showed the eCDF for absolute errors and square errors for ISONE, NYISO and PJM datasets.
It can be seen from the figures that for ISONE and NYISO, WIAE had the best overall performance for under most cases.
For PJM dataset, DeepAR and TLAE had slightly better performance.

%% file: WangTongZhao24CISS.bbl
\begin{thebibliography}{10}
\providecommand{\url}[1]{#1}
\csname url@samestyle\endcsname
\providecommand{\newblock}{\relax}
\providecommand{\bibinfo}[2]{#2}
\providecommand{\BIBentrySTDinterwordspacing}{\spaceskip=0pt\relax}
\providecommand{\BIBentryALTinterwordstretchfactor}{4}
\providecommand{\BIBentryALTinterwordspacing}{\spaceskip=\fontdimen2\font plus
\BIBentryALTinterwordstretchfactor\fontdimen3\font minus \fontdimen4\font\relax}
\providecommand{\BIBforeignlanguage}[2]{{%
\expandafter\ifx\csname l@#1\endcsname\relax
\typeout{** WARNING: IEEEtran.bst: No hyphenation pattern has been}%
\typeout{** loaded for the language `#1'. Using the pattern for}%
\typeout{** the default language instead.}%
\else
\language=\csname l@#1\endcsname
\fi
#2}}
\providecommand{\BIBdecl}{\relax}
\BIBdecl

\bibitem{weron_forecasting_2008}
\BIBentryALTinterwordspacing
R.~Weron and A.~Misiorek, ``Forecasting spot electricity prices: {A} comparison of parametric and semiparametric time series models,'' \emph{International Journal of Forecasting}, vol.~24, no.~4, pp. 744--763, 2008. [Online]. Available: \url{https://www.sciencedirect.com/science/article/pii/S0169207008000952}
\BIBentrySTDinterwordspacing

\bibitem{weron_electricity_2014}
\BIBentryALTinterwordspacing
R.~Weron, ``\BIBforeignlanguage{en}{Electricity price forecasting: {A} review of the state-of-the-art with a look into the future},'' \emph{\BIBforeignlanguage{en}{International Journal of Forecasting}}, vol.~30, no.~4, pp. 1030--1081, Oct. 2014. [Online]. Available: \url{https://www.sciencedirect.com/science/article/pii/S0169207014001083}
\BIBentrySTDinterwordspacing

\bibitem{hardle_review_1997}
\BIBentryALTinterwordspacing
W.~Härdle, H.~Lütkepohl, and R.~Chen, ``A {review} of {nonparametric} {time} {series} {analysis},'' \emph{International Statistical Review / Revue Internationale de Statistique}, vol.~65, no.~1, pp. 49--72, 1997, publisher: [Wiley, International Statistical Institute (ISI)]. [Online]. Available: \url{https://www.jstor.org/stable/1403432}
\BIBentrySTDinterwordspacing

\bibitem{Wiener:58Book}
N.~Wiener, \emph{Nonlinear Problems in Random Theory}.\hskip 1em plus 0.5em minus 0.4em\relax Cambridge, MA: Technology Press of Massachusetts Institute of Technology, 1958.

\bibitem{Rosenblatt:59}
M.~Rosenblatt, ``Stationary processes as shifts of functions of independent random variables,'' \emph{{Journal of Mathematics and Mechanics}}, vol.~8, no.~5, pp. 665--681, 1959.

\bibitem{Kailath1968TAC}
T.~{Kailath}, ``An innovations approach to least-squares estimation--{P}art {I}: Linear filtering in additive white noise,'' \emph{IEEE Transactions on Automatic Control}, vol.~13, no.~6, pp. 646--655, 1968.

\bibitem{review_nonparametric}
A.~Gautam and V.~Singh, ``Parametric versus non-parametric time series forecasting methods: A review,'' \emph{Journal of Engineering Science and Technology Review}, vol.~13, pp. 165--171, 2020.

\bibitem{nguyen_temporal_2021}
\BIBentryALTinterwordspacing
N.~Nguyen and B.~Quanz, ``\BIBforeignlanguage{en}{Temporal {latent} {auto}-{encoder}: {a} {method} for {probabilistic} {multivariate} {time} {series} {forecasting}},'' \emph{\BIBforeignlanguage{en}{Proceedings of the AAAI Conference on Artificial Intelligence}}, vol.~35, no.~10, pp. 9117--9125, May 2021, number: 10. [Online]. Available: \url{https://ojs.aaai.org/index.php/AAAI/article/view/17101}
\BIBentrySTDinterwordspacing

\bibitem{li_synergetic_2021}
\BIBentryALTinterwordspacing
L.~Li, J.~Zhang, J.~Yan, Y.~Jin, Y.~Zhang, Y.~Duan, and G.~Tian, ``\BIBforeignlanguage{en}{Synergetic {learning} of {heterogeneous} {temporal} {sequences} for {multi}-{horizon} {probabilistic} {forecasting}},'' \emph{\BIBforeignlanguage{en}{Proceedings of the AAAI Conference on Artificial Intelligence}}, vol.~35, no.~10, pp. 8420--8428, May 2021, number: 10. [Online]. Available: \url{https://ojs.aaai.org/index.php/AAAI/article/view/17023}
\BIBentrySTDinterwordspacing

\bibitem{rasul_multivariate_2022}
\BIBentryALTinterwordspacing
K.~Rasul, A.-S. Sheikh, I.~Schuster, U.~M. Bergmann, and R.~Vollgraf, ``\BIBforeignlanguage{en}{Multivariate {probabilistic} {time} {series} {forecasting} via {conditioned} {normalizing} {flows}},'' Feb. 2022. [Online]. Available: \url{https://openreview.net/forum?id=WiGQBFuVRv}
\BIBentrySTDinterwordspacing

\bibitem{Li_2022_diffusion}
\BIBentryALTinterwordspacing
Y.~Li, X.~Lu, Y.~Wang, and D.~Dou, ``Generative time series forecasting with diffusion, denoise, and disentanglement,'' in \emph{Advances in Neural Information Processing Systems}, S.~Koyejo, S.~Mohamed, A.~Agarwal, D.~Belgrave, K.~Cho, and A.~Oh, Eds., vol.~35.\hskip 1em plus 0.5em minus 0.4em\relax Curran Associates, Inc., 2022, pp. 23\,009--23\,022. [Online]. Available: \url{https://proceedings.neurips.cc/paper_files/paper/2022/file/91a85f3fb8f570e6be52b333b5ab017a-Paper-Conference.pdf}
\BIBentrySTDinterwordspacing

\bibitem{salinas_deepar_2019}
\BIBentryALTinterwordspacing
D.~Salinas, V.~Flunkert, and J.~Gasthaus, ``{DeepAR}: {probabilistic} {forecasting} with {autoregressive} {recurrent} {networks},'' Feb. 2019, arXiv:1704.04110 [cs, stat]. [Online]. Available: \url{http://arxiv.org/abs/1704.04110}
\BIBentrySTDinterwordspacing

\bibitem{wang_deep_2019}
\BIBentryALTinterwordspacing
Y.~Wang, A.~Smola, D.~C. Maddix, J.~Gasthaus, D.~Foster, and T.~Januschowski, ``Deep {factors} for {forecasting},'' May 2019, arXiv:1905.12417 [cs, stat]. [Online]. Available: \url{http://arxiv.org/abs/1905.12417}
\BIBentrySTDinterwordspacing

\bibitem{du_probabilistic_2022}
\BIBentryALTinterwordspacing
H.~Du, S.~Du, and W.~Li, ``\BIBforeignlanguage{en}{Probabilistic time series forecasting with deep non-linear state space models},'' \emph{\BIBforeignlanguage{en}{CAAI Transactions on Intelligence Technology}}, vol. n/a, no. n/a, 2022, \_eprint: https://ietresearch.onlinelibrary.wiley.com/doi/pdf/10.1049/cit2.12085. [Online]. Available: \url{https://onlinelibrary.wiley.com/doi/abs/10.1049/cit2.12085}
\BIBentrySTDinterwordspacing

\bibitem{oord_wavenet_2016}
\BIBentryALTinterwordspacing
A.~v.~d. Oord, S.~Dieleman, H.~Zen, K.~Simonyan, O.~Vinyals, A.~Graves, N.~Kalchbrenner, A.~Senior, and K.~Kavukcuoglu, ``{WaveNet}: {a} {generative} {model} for {raw} {audio},'' Sep. 2016, arXiv:1609.03499 [cs]. [Online]. Available: \url{http://arxiv.org/abs/1609.03499}
\BIBentrySTDinterwordspacing

\bibitem{borovykh_conditional_2018}
\BIBentryALTinterwordspacing
A.~Borovykh, S.~Bohte, and C.~W. Oosterlee, ``Conditional {time} {series} {forecasting} with {convolutional} {neural} {networks},'' Sep. 2018, arXiv:1703.04691 [stat]. [Online]. Available: \url{http://arxiv.org/abs/1703.04691}
\BIBentrySTDinterwordspacing

\bibitem{zhou_informer_2021}
\BIBentryALTinterwordspacing
H.~Zhou, S.~Zhang, J.~Peng, S.~Zhang, J.~Li, H.~Xiong, and W.~Zhang, ``\BIBforeignlanguage{en}{Informer: {beyond} {efficient} {transformer} for {long} {sequence} {time}-{series} {forecasting}},'' \emph{\BIBforeignlanguage{en}{Proceedings of the AAAI Conference on Artificial Intelligence}}, vol.~35, no.~12, pp. 11\,106--11\,115, May 2021, number: 12. [Online]. Available: \url{https://ojs.aaai.org/index.php/AAAI/article/view/17325}
\BIBentrySTDinterwordspacing

\bibitem{zhou_fedformer_2022}
\BIBentryALTinterwordspacing
T.~Zhou, Z.~Ma, Q.~Wen, X.~Wang, L.~Sun, and R.~Jin, ``{FEDformer}: {frequency} {enhanced} {decomposed} {transformer} for {long}-term {series} {forecasting},'' Jun. 2022, arXiv:2201.12740 [cs, stat]. [Online]. Available: \url{http://arxiv.org/abs/2201.12740}
\BIBentrySTDinterwordspacing

\bibitem{liu_pyraformer_2022}
\BIBentryALTinterwordspacing
S.~Liu, H.~Yu, C.~Liao, J.~Li, W.~Lin, A.~X. Liu, and S.~Dustdar, ``\BIBforeignlanguage{en}{Pyraformer: {low}-{complexity} {pyramidal} {attention} for {long}-{range} {time} {series} {modeling} and {forecasting}},'' Mar. 2022. [Online]. Available: \url{https://openreview.net/forum?id=0EXmFzUn5I}
\BIBentrySTDinterwordspacing

\bibitem{Kalman:60TASME}
R.~E. Kalman, ``A new approach to linear filtering and prediction problems,'' \emph{{Trans. ASME J. of Basic Engineering}}, vol.~82, no.~1, pp. 35--45, 1960.

\bibitem{WangTong:21JMLR}
\BIBentryALTinterwordspacing
X.~Wang and L.~Tong, ``Innovations autoencoder and its application in one-class anomalous sequence detection,'' \emph{Journal of Machine Learning Research}, vol.~23, no.~49, pp. 1--27, 2022. [Online]. Available: \url{http://jmlr.org/papers/v23/21-0735.html}
\BIBentrySTDinterwordspacing

\bibitem{BickelDoksum07:book}
P.~J. Bickel and K.~A. Doksum, \emph{Mathematical statistics: basic ideas and selected topics. (2nd ed.)}.\hskip 1em plus 0.5em minus 0.4em\relax Upper Saddle River, N.J.: Pearson Prentice Hall, 2007, vol.~1.

\bibitem{wang2023nonparametric}
X.~Wang, M.~Lee, Q.~Zhao, and L.~Tong, ``Non-parametric probabilistic time series forecasting via innovations representation,'' \emph{arXiv preprint arXiv:2306.03782}, 2023.

\bibitem{Arjovsky17}
M.~Arjovsky, S.~Chintala, and L.Bottou, ``{Wasserstein GAN},'' Jan. 2017, arXiv:1701.07875.

\bibitem{alexandrov_gluonts_2019}
\BIBentryALTinterwordspacing
A.~Alexandrov, K.~Benidis, M.~Bohlke-Schneider, V.~Flunkert, J.~Gasthaus, T.~Januschowski, D.~C. Maddix, S.~Rangapuram, D.~Salinas, J.~Schulz, L.~Stella, A.~C. Türkmen, and Y.~Wang, ``{GluonTS}: {probabilistic} {time} {series} {models} in {Python},'' Jun. 2019, arXiv:1906.05264 [cs, stat]. [Online]. Available: \url{http://arxiv.org/abs/1906.05264}
\BIBentrySTDinterwordspacing

\bibitem{Ji&Thomas&Tong:17TPS}
Y.~Ji, R.~J. Thomas, and L.~Tong, ``Probabilistic forecasting of real-time lmp and network congestion,'' \emph{IEEE Transactions on Power Systems}, vol.~32, no.~2, pp. 831--841, 2017.

\bibitem{Ji&Deng&Tong:21Bkchap}
Y.~Ji, L.~Tong, and W.~Deng, ``Probabilistic forecasting of power system and market operations,'' in \emph{{Advanced data analytics for power systems}}.\hskip 1em plus 0.5em minus 0.4em\relax Cambridge : Cambridge University Press, 2021.

\end{thebibliography}
